\documentclass{article}

\usepackage{microtype}
\usepackage{graphicx}
\usepackage{subfigure}
\usepackage{booktabs} 

\usepackage{hyperref}

\usepackage{hyperref}
\usepackage{url}

\usepackage{array}
\usepackage{color}

\usepackage{multirow}
\usepackage{amssymb}
\usepackage{bm}
\usepackage{balance}
\usepackage{arydshln}
\usepackage{mathtools}
\usepackage{soul}
\usepackage[algo2e]{algorithm2e}
\usepackage{enumitem}
\usepackage{caption}
\usepackage{wrapfig}

\usepackage{booktabs}
\usepackage{amsmath,amsfonts}
\usepackage{xspace}

\usepackage{diagbox}

\usepackage{bbm}

\usepackage{xspace}
\usepackage{textcomp}

\usepackage{tabularx}
\usepackage{verbatim}
\usepackage{tablefootnote}

\newcommand{\yong}[1]{{\color{blue}{(Yong\@: #1)}}}
\newcommand{\fancyname}{\emph{$F_2$-XEnDec}}

\def\max{\mathop{\rm max}}%

\def\S{\mathcal{S}} %

\def\vx{\mathbf{x}}
\def\vy{\mathbf{y}}
\def\vz{\mathbf{z}}
\def\vc{\mathbf{c}}
\def\vm{\mathbf{m}}
\def\mA{\mathbf{A}}

\def\bx{\vx} %
\def\bxp{\vx^{\prime}} %
\def\vxp{\vx^{\prime}} %
\def\xp{x^{\prime}} %
\def\tbx{\tilde{\vx}} %
\def\tvx{\tilde{\vx}} %

\def\by{\vy} %
\def\byp{\vy^{\prime}} %
\def\vyp{\vy^{\prime}} %
\def\tby{\tilde{\vy}} %
\def\tvy{\tilde{\vy}} %

\def\tvz{\tilde{\vz}} %

\newcommand{\eat}[1]{} 
\newcommand{\eg}{\emph{e.g.}} 
\newcommand{\ie}{\emph{i.e.}}

\usepackage[accepted]{icml2021}

\icmltitlerunning{Self-supervised and Supervised Joint Training for Resource-rich Machine Translation}

\begin{document}

\twocolumn[
\icmltitle{Self-supervised and Supervised Joint Training for \\ Resource-rich Machine Translation}



\icmlsetsymbol{note}{$^\dagger$}

\begin{icmlauthorlist}
\icmlauthor{Yong Cheng}{goo}
\icmlauthor{Wei Wang}{note}
\icmlauthor{Lu Jiang}{goo,cmu}
\icmlauthor{Wolfgang Macherey}{goo}
\end{icmlauthorlist}

\icmlaffiliation{goo}{Google Research, Google LLC, USA}
\icmlaffiliation{cmu}{Language Technologies Institute, Carnegie Mellon University, Pittsburgh, Pennsylvania}

\icmlcorrespondingauthor{Yong Cheng}{chengyong@google.com}

\icmlkeywords{Machine Learning, ICML}

\vskip 0.3in
]


\printAffiliationsAndNotice{}  

\begin{abstract}
Self-supervised pre-training of text representations has been successfully applied to low-resource Neural Machine Translation (NMT). However, it usually fails to achieve notable gains on resource-rich NMT.
In this paper, we propose a joint training approach, \fancyname, to combine self-supervised and supervised learning to optimize NMT models. To exploit complementary self-supervised signals for supervised learning, NMT models are trained on examples that are interbred from monolingual and parallel sentences through a new process called crossover encoder-decoder.
Experiments on two resource-rich translation benchmarks, WMT'14 English-German and WMT'14 English-French, demonstrate that our approach achieves substantial improvements over several strong baseline methods and obtains a new state of the art of 46.19 BLEU on English-French when incorporating back translation. Results also show that our approach is capable of improving model robustness to input perturbations such as code-switching noise which frequently appears on social media.
\end{abstract}

\section{Introduction}
Self-supervised pre-training of text
representations~\citep{peters2018deep,radford2018improving} has achieved tremendous success in natural language processing applications.
Inspired by BERT~\citep{devlin2019bert}, recent works attempt to leverage sequence-to-sequence model pre-training for Neural Machine Translation (NMT)~\citep{lewis2019bart,song2019mass,liu2020multilingual}.
Generally, these methods comprise two stages: pre-training and finetuning.
During the pre-training stage, the model is learned with a self-supervised task on abundant unlabeled data (\ie~monolingual sentences). In the second stage, the full or partial model is finetuned on a downstream translation task of labeled data (\ie~parallel sentences).
Studies have demonstrated the benefit of pre-training for the low-resource translation task in which the labeled data is limited~\citep{lewis2019bart,song2019mass}. All these successes share the same setup: pre-training on abundant unlabeled data and finetuning on limited labeled data.


In many NMT applications, we are confronted with a different setup where abundant labeled data, \eg, millions of parallel sentences, are available for finetuning. For these resource-rich translation tasks, the two-stage approach is less effective and, even worse, sometimes can undermine the performance if improperly utilized~\citep{zhu2020bertnmt}, in part due to the catastrophic forgetting~\citep{french1999catastrophic}. More recently, several mitigation techniques have been proposed for the two-stage approach~\citep{edunov2019pre,yang2019towards,zhu2020bertnmt}, such as freezing the pre-trained representations during finetuning. However, these strategies hinder uncovering the full potential of self-supervised learning since the learned representations are either held fixed or slightly tuned in the supervised learning.

In this paper, we study resource-rich machine translation through a different perspective of joint training where, in contrast to the conventional two-stage approaches, we train NMT models in a single stage using the self-supervised objective (on monolingual sentences) in addition to the supervised objective (on parallel sentences).
The challenge for this single-stage training paradigm is that self-supervised learning is
less useful in joint training because it provides a much weaker learning signal that can be easily dominated by the supervised learning signal in joint training. As a result,
conventional approaches such as combining self-supervised and supervised learning objectives perform not much better than the supervised learning objective by itself.

This paper aims at exploiting the complementary signals in self-supervised learning to facilitate supervised learning. Inspired by chromosomal crossovers~\citep{rieger2012glossary}, we propose an essential new task called crossover encoder-decoder (or \emph{XEnDec}) which takes two training examples as inputs (called parents), shuffles their source sentences, and produces a sentence by a mixture decoder model. Our method applies \emph{XEnDec} to ``deeply'' fuse the monolingual (unlabeled) and parallel (labeled) sentences, thereby producing their first and second filial generation (or $F_1$ and $F_2$ generation). As we find that the $F_2$ generation exhibits combinations of traits that differ from those found in the monolingual or the parallel sentence, we train NMT models on the $F_2$ offspring and name our method \emph{$F_2$-XEnDec}.

To the best of our knowledge, the proposed method is among the first NMT models on joint self-supervised and supervised learning, and moreover, the first to demonstrate such joint learning substantially benefits resource-rich machine translation. Compared to recent two-stage finetuning approaches \cite{zhu2020bertnmt} and \cite{yang2019towards}, our method only needs a single training stage to utilize the complementary signals in self-supervised learning. Empirically, our results show the proposed single-stage approach achieves comparable or better results than previous methods. In addition, our method improves the robustness of NMT models which is known as a critical deficiency in contemporary NMT systems (cf.~Section~\ref{sec:exp_analysis}). It is noteworthy that none of the two-stage training approaches have ever reported this behavior.

We empirically validate our approach on the WMT'14 English-German and WMT'14 English-French translation benchmarks which yields an improvement of $2.13$ and $1.78$ BLEU points over the vanilla Transformer model~\cite{ott2018scaling}, respectively. It achieves a new state of the art of 46.19 BLEU on the WMT'14 English-French translation task with the back translation technique.
In summary, our contributions are as follows:
\begin{enumerate}
    \item We propose a crossover encoder-decoder (\emph{XEnDec}) which, with appropriate inputs, can reproduce several existing self-supervised and supervised learning objectives.
    \item We jointly train self-supervised and supervised objectives in a single stage, and show that our method is able to exploit the complementary signals in self-supervised learning to facilitate supervised learning.
    \item Our approach achieves significant improvements on resource-rich translation tasks and exhibits higher robustness against input perturbations such as code-switching noise.
\end{enumerate}


\begin{figure*}[!t]
\begin{center}
\subfigure{\label{figure:approach_a}}
\subfigure{\label{figure:approach_b}}
\includegraphics[width=0.98\textwidth]{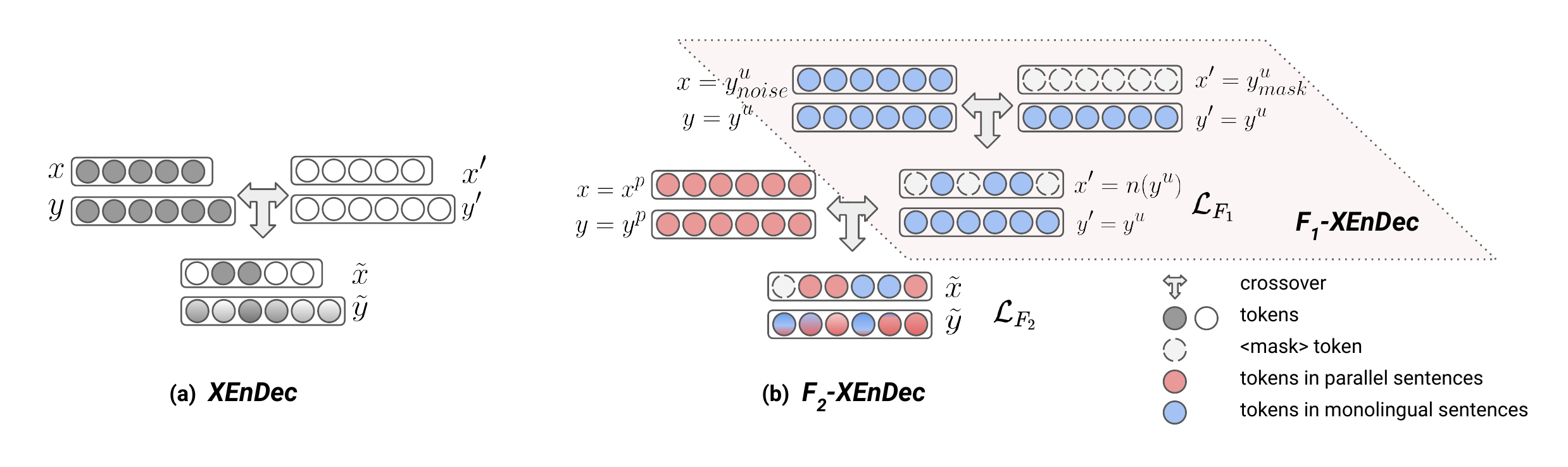}
\caption{\textbf{(a)} Illustration of crossover encoder-decoder (\emph{XEnDec}). It takes two training examples  $(\vx, \vy)$ and $(\vxp, \vyp)$ as inputs, and outputs a sentence pair ($\tvx$, $\tvy$). \textbf{(b)} Our method applies \emph{XEnDec} to fuse the monolingual (blue) and parallel sentences (red). In the first generation, \emph{$F_{1}$-XEnDec} generates $(n(\vy^u), \vy^u)$ incurring a self-supervised loss $\mathcal{L}_{F_1}$, where $(n(\vy^u)$ is the function discussed in Section~\ref{sec:pretraining_nmt} that corrupts the monolingual sentence $\vy^u$. \emph{$F_{2}$-XEnDec} applies another round of \emph{XEnDec} to incorporate parallel data $(\vx^p, \vy^p)$ to get the $F_{2}$ output $(\tvx, \tvy)$.
$\vy^{u}$: a monolingual sentence.
$\vy^{u}_{noise}$: a sentence generated by adding non-masking noise to $\vy^{u}$.  $\vy^{u}_{mask}$: a sentence of length $|\vy^{u}|$ containing only  ``$\langle mask \rangle$'' tokens.}\label{figure:approach}
\end{center}
\end{figure*}
\section{Background}

\subsection{Neural Machine Translation}\label{sec:bg_nmt}
%
Under the encoder-decoder paradigm ~\citep{Bahdanau:15, Gehring:17, Vaswani:17}, the conditional probability $P(\vy|\vx;\bm{\theta})$ of a target-language sentence $\vy = y_{1}, \cdots, y_{J}$ given a source-language sentence $\vx = x_{1},\cdots, x_{I}$ is modeled as follows:
The encoder maps the source sentence $\vx$ onto a sequence of $I$ word embeddings $e(\vx) = e(x_{1}),...,e(x_{I})$. Then the word embeddings are encoded into their corresponding continuous hidden representations. The decoder acts as a conditional language model that reads embeddings $e(\vy)$ for a shifted copy of $\vy$ along with the aggregated contextual representations $\vc$.
For clarity, we denote the input and output in the decoder as $\vz$ and $\vy$, \ie, $\vz = \langle s \rangle, y_{1}, \cdots, y_{J-1}$, where $\langle s \rangle$ is a start symbol.
Conditioned on an aggregated contextual representation $\vc_j$ and its partial target input $\vz_{\leq j}$, the decoder generates $\vy$ as:
\begin{eqnarray}
P(\vy|\vx;\bm{\theta}) = \prod_{j=1}^{J} P(y_{j}|\vz_{\leq j}, \vc;\bm{\theta}).
\end{eqnarray}
The aggregated contextual representation $\vc$ is often calculated by summarizing the sentence $\vx$ with an attention mechanism~\citep{Bahdanau:15}. A byproduct of the attention computation is a noisy alignment matrix $\mA \in \mathbb{R}^{J \times I}$ which roughly captures the translation correspondence between target and source words~\citep{garg2019jointly}. 

Generally, NMT optimizes the model parameters $\bm{\theta}$ by minimizing the empirical risk over a parallel training set $(\vx, \vy) \in \mathcal{S}$:
\begin{eqnarray}
\mathcal{L}_{\mathcal{S}}(\bm{\theta}) = \mathop{\mathbb{E}}\limits_{(\vx, \vy) \in \mathcal{S}} \lbrack \ell(f(\vx, \vy;\bm{\theta}), h(\vy)) \rbrack,
\label{eq:loss_clean}
\end{eqnarray}
where $\ell$ is the cross entropy loss between the model prediction $f(\vx, \vy;\bm{\theta})$ and $h(\vy)$, and $h(\vy)$ denotes the sequence of one-hot label vectors 
 with label smoothing in the Transformer~\citep{Vaswani:17}.

\subsection{Pre-training for Neural Machine Translation}\label{sec:pretraining_nmt}
Pre-training sequence-to-sequence models for language generation has been shown to be effective for machine translation~\citep{song2019mass,lewis2019bart}. These methods generally comprise two stages: pre-training and finetuning. The pre-training takes advantage of an abundant monolingual corpus $\mathcal{U} = \{ \vy\}$ to learn representations through a self-supervised objective called denoising autoencoder~\citep{vincent2008extracting} which aims at reconstructing the original sentence $\by$ from one of its corrupted counterparts.

Let $n(\vy)$ be a corrupted copy of $\by$ where the function $n(\cdot)$ adds noise and/or masks words.
$(n(\vy), \vy)$ constitutes the pseudo parallel data and is fed into the NMT model to compute the reconstruction loss. The self-supervised reconstruction loss over the corpus $\mathcal{U}$ is defined as:
\begin{eqnarray}
\mathcal{L}_{\mathcal{U}}(\bm{\theta}) = \mathop{\mathbb{E}}\limits_{\vy \in \mathcal{U}} \lbrack \ell(f(n(\vy), \vy;\bm{\theta}), h(\vy)) \rbrack,
\label{eq:loss_self}
\end{eqnarray}
The optimal model parameters $\bm{\theta}^{\star}$ are learned via the self-supervised loss
$\mathcal{L}_{\mathcal{U}}(\bm{\theta})$ and used to initialize downstream models during the finetuning on the parallel training set $\S$. 


\section{Cross-breeding: \fancyname}
For resource-rich translation tasks in which a large parallel corpus and (virtually) unlimited monolingual corpora are available, our goal is to improve translation performance by exploiting self-supervised signals to complement the supervised learning. In the proposed method, we train NMT models jointly with supervised and self-supervised learning objectives in a single stage. This is based on an essential task called \emph{XEnDec}. In the remainder of this section, we first detail the \emph{XEnDec} and then introduce our approach and present the overall algorithm. Finally, we discuss its relationship to some of the previous works.


\subsection{Crossover Encoder-Decoder}
This section introduces the crossover encoder-decoder (\emph{XEnDec}). Different from a conventional encoder-decoder, \emph{XEnDec} takes two training examples as inputs (called parents), shuffles the parents' source sentences and produces a virtual example (called offspring) through a mixture decoder model. Fig.~\ref{figure:approach_a} illustrates this process.

Formally, let $(\vx, \vy)$ denote a training example where $\vx=x_1, \cdots, x_I$ represents a source sentence of $I$ words and $\vy=y_1, \cdots, y_{J}$ is the corresponding target sentence of $J$ words. In supervised training, $\vx$ and $\vy$ are parallel sentences. As we will see in Section~\ref{sec:our_algorithm}, \emph{XEnDec} can be carried out with and without supervision. We do not distinguish these cases for now and use generic notations to illustrate the idea.

Given a pair of examples $(\vx, \by)$ and $(\vxp, \vyp)$ called parents, the crossover encoder shuffles the two source sequences into a new source sentence $\tvx$, calculated from:
\begin{eqnarray}
\label{eq:shuffled_source}
\tilde{x}_{i} = m_{i} x_i + (1 - m_{i}) \xp_{i}, \label{eq:mix_src}
\end{eqnarray}
where $\vm = m_1, \cdots, m_I \in \{0,1\}^I$ stands for a series of Bernoulli random variables with each taking the value 1 with probability $p$ called shuffling ratio.
If $m_i=0$, then the $i$-th word in $\bx$ will be substituted with the word in $\bxp$ at the same position. For convenience, the lengths of the two sequences are aligned by appending padding tokens to the end of the shorter sentence. 

The crossover decoder employs a mixture model to generate the virtual target sentence. The embedding of the decoder's input $\tvz$ is computed as:
\begin{equation}
\begin{split}
e(\tilde{z}_{j}) =& \frac{1}{Z} \big[ e(y_{j-1}) \sum_{i=1}^I A_{(j-1)i}m_i\\
&+ e(y^{\prime}_{j-1}) \sum_{i=1}^I A'_{(j-1)i}(1-m_i) \big], \label{eq:mix_tgt}
\end{split}
\end{equation}
where $e(\cdot)$ is the embedding function.
$Z=\sum_{i=1}^IA_{(j-1)i}m_i + A'_{(j-1)i}(1-m_i)$ is the normalization term where $\mA$ and $\mA'$ are the alignment matrices for the source sequences $\bx$ and $\bxp$, respectively. 
Eq.~\eqref{eq:mix_tgt} averages embeddings of $\vy$ and $\vy^{\prime}$ through the latent weights computed by $\vm$, $\mA$, and $\mA'$.
The alignment matrix measures the contribution of the source words for generating a specific target word~\citep{och2004alignment,Bahdanau:15}. For example, 
$A_{ji}$ represents the contribution score of the $i$-th word in the source sentence for the $j$-th word in the target sentence. For simplicity, this paper uses the attention matrix learned in the NMT model as a noisy alignment matrix~\citep{garg2019jointly}.

Likewise, the label vector for the crossover decoder is calculated from:
\begin{equation}
\begin{split}
h(\tilde{y}_{j}) =&\frac{1}{Z} \big[ h(y_j) \sum_{i=1}^I A_{ji}m_i  \\
&+ h(y^{\prime}_{j}) \sum_{i=1}^I A'_{ji}(1-m_i) \big],\label{eq:mix_signal}
\end{split}
\end{equation}
The $h(\cdot)$ function projects a word onto its label vector, \eg, a one-hot vector.
The loss of \emph{XEnDec} is computed over its output ($\tvx$, $\tvy$) using the negative log-likelihood:
\begin{equation}
\begin{split}
&\quad\ell (f(\tvx, \tvy; \bm{\theta}), h(\tvy)) = -\log P(\tby|\tbx;\bm{\theta}) \\ 
&=\sum_{j} KL(h(\tilde{y}_{j}) \| P(y|\tvz_{\leq j}, \vc_{j};\bm{\theta})),\label{eq:crossover_loss}
\end{split}
\end{equation}
where $\tvz$ is a shifted copy of $\tvy$ as discussed in Section~\ref{sec:bg_nmt}. Notice that even though we do not directly observe the ``virtual sentences'' $\tvz$ and $\tvy$, we are still able to compute the loss using their embeddings and labels. In practice, the length of $\tilde{\vx}$ is set to $\max(|\vx|, |\vx^{\prime}|)$ whereas $\tilde{\vy}$ and $\tilde{\vz}$ share the same length of $\max(|\vy|, |\vy^{\prime}|)$.

\eat{
\subsection{Discussion}
\yong{Why does it work?
1. It designs a new training task.
2. It fills the gap between two training objectives.
3. As a strong regularization.}

The core term $\mathcal{L}_{F_{2}}$ proposed in our training objective (Eq.~\eqref{eq:loss_all}) aims to combine parallel and monolingual training examples respectively used in supervised and self-supervised objectives.
We replace partial words of an input sentence in the supervised task with words from the input sentence in the self-supervised task.
The training signals are also combined proportionally to the importance of words in the combined sentences measured by their respective alignment matrics.
The new task needs to decouple the integration of two specific tasks while predicting the soft entire target sentence based on its partial source sentence. 

Over simply jointly training $\mathcal{L}_{\mathcal{S}}$ and $\mathcal{L}_{F_{1}}$ which mechanically sums up the losses across instances, $\mathcal{L}_{F_{2}}$ in the new task further enhances the combination by deeply fusing training examples from supervised and self-supervised tasks at instance level. It bridges the gap between these two heterogeneous representation space by encouraging the self-supervised task to extract more informative representations tailored for the supervised task.

As an effective regularization term, $\mathcal{L}_{F_{2}}$ corrupts a clean sentence with words taken from another sentence to construct a virtual example. It retains the advantages of other approaches which introduce stochastic or adversarial noise to regularize the model parameters towards a more robust model~\citep{Miyato:17,Cheng:19}. However, a partial view of another natural sentence is integrated as noise in $\mathcal{L}_{F_{2}}$ which achieves better translation performance and robustness evidenced in the experiment section.
}

\subsection{Training}\label{sec:our_algorithm}

The proposed method applies \emph{XEnDec} to deeply fuse the parallel data $\mathcal{S}$ with nonparallel, monolingual data $\mathcal{U}$. As illustrated in Fig.~\ref{figure:approach_b}, the first generation (\emph{$F_1$-XEnDec} in the figure) uses \emph{XEnDec} to combine monolingual sentences of different views, thereby incurring a self-supervised loss $\mathcal{L}_{F_{1}}$. We compute the loss $\mathcal{L}_{F_{1}}$ using Eq.~\eqref{eq:loss_self}. Afterward, the second generation (\fancyname~in the figure) applies \emph{XEnDec} to the offspring of the first generation ($n(\vy^u)$, $\vy^u$) and a sampled parallel sentence ($\vx^{p}$, $\vy^{p}$), yielding a new loss term $\mathcal{L}_{F_{2}}$. The loss $\mathcal{L}_{F_{2}}$ is computed over the output of the \emph{$F_2$-XEnDec} by:
\begin{eqnarray}
\mathcal{L}_{F_{2}}(\bm{\theta}) = \mathop{\mathbb{E}}\limits_{\by^u \in \mathcal{U}} \mathop{\mathbb{E}}\limits_{(\vx^{p}, \vy^{p}) \in \mathcal{S}} \lbrack \ell(f(\tbx, \tby;\bm{\theta}), h(\tvy)) \rbrack,
\label{eq:loss_virtual}
\end{eqnarray}
where $(\tbx, \tby)$ is the output of the \emph{$F_2$-XEnDec} in Fig.~\ref{figure:approach_b}.

The final NMT models are optimized jointly on the original translation loss and the above two auxiliary losses. 
\begin{eqnarray}
\mathcal{L}({\bm{\theta}}) = \mathcal{L}_{\mathcal{S}}({\bm{\theta}}) + \mathcal{L}_{F_1}({\bm{\theta}}) +  \mathcal{L}_{F_2}(\bm{\theta}), \label{eq:loss_all}
\end{eqnarray}
$\mathcal{L}_{F_{2}}$ in Eq.~\eqref{eq:loss_all} is used to deeply fuse monolingual and parallel sentences at instance level rather than combine them mechanically.
Section~\ref{sec:ablation} empirically verifies the contributions of the $\mathcal{L}_{F_{1}}$ and $\mathcal{L}_{F_{2}}$ loss terms.

Algorithm~\ref{algo1} delineates the procedure to compute the final loss $\mathcal{L}(\bm{\theta})$. 
Specifically, each time, we sample a monolingual sentence for each parallel sentence to circumvent the expensive enumeration in Eq.~\eqref{eq:loss_virtual}. To speed up the training, we group sentences offline by length in Step 3 (cf. batching data in the supplementary document). For adding noise in Step 4, we can follow~\cite{lample2017unsupervised} to locally shuffle words while keeping the distance between the original and new position not larger than $3$ or set it as a null operation.
There are two techniques to boost the final performance.

\textbf{Computing $\mA$:} The alignment matrix $\mA$ is obtained by averaging the cross-attention weights across all decoder layers and heads.
We also add a temperature to control the sharpness of the attention distribution, the reciprocal of which was linearly increased from $0$ to $2$ during the first $20K$ steps.
To avoid overfitting when computing $e(\tvz)$ and $h(\tvy)$, 
we apply dropout to $\mA$ and stop back-propagating gradients through $\mA$ when calculating the loss $\mathcal{L}_{F_2}(\bm{\theta})$. 

\textbf{Computing $h(\tvy)$:} Instead of interpolating one-hot labels in Eq.~\eqref{eq:mix_signal}, we use the prediction vector $f(\vx, \vy;\hat{\bm{\theta}})$ on the sentence pair $(\vx, \vy)$ estimated by the model where $\hat{\bm{\theta}}$ indicates no gradients are back-propagated through it. 
However, the predictions made at early stages are usually unreliable.
We propose to linearly combine the ground-truth one-hot label with the model prediction using a parameter $v$, which is computed as $v f_{j}(\bx, \by;\hat{\bm{\theta}}) + (1 - v)h(y_{j})$
where $v$ is gradually annealed from $0$ to $1$ during the first $20K$ steps \footnote{These two annealing hyperparameters in computing both $\mA$ and $h(\tvy)$ are the same for all the models and not elaborately tuned.}. Notice that the prediction vectors are not used in computing the decoder input $e(\tvz)$ which can be clearly distinguished from schedule sampling~\cite{bengio2015scheduled}.

\begin{algorithm}[!t]
\SetAlgoLined
\LinesNumbered
\small
\KwIn{Parallel corpus $\mathcal{S}$, Monolingual corpus $\mathcal{U}$, and Shuffling ratios $p_1$ and $p_2$}
\KwOut{Batch Loss $\mathcal{L}(\bm{\theta})$.}
\SetKwFunction{algo}{\fancyname}
\SetKwProg{Fn}{Function}{:}{}
\Fn{\algo{$\mathcal{S}, \mathcal{U}, p_{1}, p_{2}$}}{
   \ForEach{$(\vx^{p}, \vy^{p}) \in \mathcal{S}$ }{
   Sample a $\vy^u \in \mathcal{U}$ with similar length as $\vx^{p}$; {\footnotesize$\;\;$\tcp{done offline.}}
   $\vy^{u}_{noise}$ $\leftarrow$ add non-masking noise to $\vy^u$;
   
    $(n(\vy^u), \vy^u)$ $\leftarrow$ \emph{XEnDec} over the inputs $(\vy^{u}_{noise}, \vy^u)$ and $(\vy^u_{mask}, \vy^u)$, with the shuffling ratio $p_{1}$ and arbitrary alignment matrices;
   
   $\mathcal{L}_\S$ $\leftarrow$ compute $\ell$ in Eq.~\eqref{eq:loss_clean} using $(\vx^{p}, \vy^{p})$ and obtain its attention matrix $\mA$;
   
   $\mathcal{L}_{F_1}$ $\leftarrow$ compute $\ell$ in Eq.~\eqref{eq:loss_self} using $(n(\vy^u), \vy^u)$ and obtain $\mA^{\prime}$;
   
   $(\tvx, \tvy)$ $\leftarrow$ \emph{XEnDec} over the inputs $(\vx^{p}, \vy^{p})$ and $(n(\vy^u), \vy^{u})$, with the shuffling ratio $p_{2}$, $\mA$ and $\mA^{\prime}$;

    $\mathcal{L}_{F_2}$ $\leftarrow$ compute $\ell$ in Eq.~\eqref{eq:loss_virtual};
  }
 \KwRet $\mathcal{L}(\bm{\theta}) = \mathcal{L}_{\mathcal{S}}({\bm{\theta}}) + \mathcal{L}_{F_1}({\bm{\theta}}) +  \mathcal{L}_{F_2}(\bm{\theta})$; {\footnotesize$\;\;$\tcp{Eq.~\eqref{eq:loss_all}.}}

 }
 \caption{Proposed {\em \fancyname} function.} \label{algo1}
\end{algorithm}

\subsection{Relation to Other Works}
\begin{table}[t]
\centering
\caption{Comparison with different objectives produced by \emph{XEnDec}. Each row shows a set of inputs to \emph{XEnDec} and the corresponding objectives in existing work (the last column). $\vy_{mask}^{u}$ is a sentence of length $|\vy^{u}|$ containing only ``$\langle mask \rangle$'' tokens. $\vy_{noise}^{u}$ is a sentence obtained by corrupting all the words in $\vy^{u}$ with non-masking noises.
$\vx^{p}_{adv}$ and $\vy^{p}_{adv}$ are adversarial sentences in which all the words are substituted with adversarial words.}
\vspace{2mm}
\small
\begin{tabular}{llll|l}
\toprule
($\bx$  &$\by$) &($\bxp$ &$\byp$)&\bf Objectives
\\ \midrule
$\by^{u}$         &$\by^{u}$  &$\by^{u}_{mask}$ &$\by^{u}$  &MASS ~\citep{song2019mass}\\
$\by^{u}_{noise}$        &$\by^{u}$  &$\by^{u}_{mask}$ &$\by^{u}$ &BART ~\citep{lewis2019bart} \\
$\bx^{p}$         &$\by^{p}$  &$\bx^{p}_{adv}$  &$\by^{p}_{adv}$ &Adv. ~\citep{Cheng:19} \\
\bottomrule
\end{tabular} \label{table:comparison_obj}
\end{table}

\enlargethispage*{1ex}

This subsection shows that \emph{XEnDec}, when fed with appropriate inputs, yields 
learning objectives identical to two recently proposed self-supervised learning approaches: MASS~\citep{song2019mass} and BART~\citep{lewis2019bart}, as well as a supervised learning approach called \textit{Doubly Adversarial}~\citep{Cheng:19}. Table~\ref{table:comparison_obj} summarizes the inputs of \emph{XEnDec} to recover these approaches.

\emph{XEnDec} can be used for self-supervised learning. As shown in Table~\ref{table:comparison_obj}, the inputs to \emph{XEnDec} are two pairs of sentences $(\vx, \vy)$ and $(\vxp, \vyp)$. Given arbitrary alignment matrices, 
if we set $\vxp=\vy^{u}$, $\vyp=\vy^{u}$, and $\vx$ to be a corrupted copy of $\vy^{u}$, then \emph{XEnDec} is equivalent to the denoising autoencoder which is commonly used to pre-train sequence-to-sequence models such as in MASS~\citep{song2019mass} and BART~\citep{lewis2019bart}.
In particular, if we allow $\bxp$ to be a dummy sentence of length $|\vy^{u}|$ containing only ``$\langle mask \rangle$'' tokens ($\by^{u}_{mask}$ in the table), Eq.~\eqref{eq:crossover_loss} yields the learning objective
defined in the MASS model~\citep{song2019mass} except that losses over unmasked words are not counted in the training loss. Likewise, as shown in Table~\ref{table:comparison_obj}, we can recover BART's objective by setting $\vx=\vy^{u}_{noise}$ where $\vy^{u}_{noise}$ is obtained by  shuffling tokens or dropping them in $\vy^{u}$. In both cases, \emph{XEnDec} is trained with a self-supervised objective to reconstruct the original sentence from one of its corrupted sentences. 
Conceptually, denoising autoencoder can be regarded as a degenerated \emph{XEnDec} in which the inputs are two views of its source correspondence for a monolingual sentence, \eg, $n(\vy)$ and $\vy_{mask}$ for $\vy$.

\emph{XEnDec} can also be used in supervised learning. The translation loss proposed in~\citep{Cheng:19} is achieved by letting $\vxp$ and $\vyp$ be two ``adversarial inputs'', $\bx^{p}_{adv}$ and $\by^{p}_{adv}$, both of which consist of adversarial words at each position. For the construction of $\bx^{p}_{adv}$, we refer to Algorithm~1 in~\citep{Cheng:19}. In this case, the crossover encoder-decoder is trained with a supervised objective over parallel sentences.

The above connections to existing works illustrate the power of \emph{XEnDec} when it is fed with different kinds of inputs. The results in Section~\ref{sec:ablation} show that \emph{XEnDec} is still able to improve the baseline with alternative inputs. However, our experiments show the best configuration found so far is to use the \emph{$F_2$-XEnDec} in Algorithm~\ref{algo1} to deeply fuse the monolingual and parallel sentences.

\section{Experiments}
\begin{table*}[t]
\caption{Experiments on WMT'14 English-German and WMT'14 English-French translation.}
\centering
\begin{tabular}{ll|ll|ll}
\toprule
\multicolumn{1}{l}{Models}  & \multicolumn{1}{l|}{Methods} &\multicolumn{1}{l}{En$\rightarrow$De} &\multicolumn{1}{l|}{De$\rightarrow$En}  &{En$\rightarrow$Fr} & {Fr$\rightarrow$En}
\\ \midrule
\multirow{2}{*}{Base} &Reproduced Transformer &$28.70$ &$32.23$ &- &- \\
                      &\fancyname &$\textbf{30.46}$ &$\textbf{34.06}$ &- &- \\
\midrule
\multirow{4}{*}{Big} &Reproduced Transformer &$29.47$ &$33.12$ &$43.37$ &$39.82$ \\
                      &\cite{ott2018scaling} &$29.30$ &- &$43.20$ &- \\
                      &\cite{Cheng:19} &$30.01$ &- &- &- \\
                      &\cite{yang2019towards} &$30.10$ &- &$42.30$ & \\
                      &\cite{nguyen2019data} &$30.70$ &- &$43.70$ &- \\
                      &\cite{zhu2020bertnmt} &$30.75$ &- &$43.78$ &- \\
                      &Joint Training with MASS &30.63 &- &43.00 &- \\
                      &Joint Training with BART &30.88 &- &44.18 &- \\
                      &\fancyname &$\textbf{31.60}$ &$\textbf{34.94}$ &$\textbf{45.15}$ &$\textbf{41.60}$ \\
\bottomrule
\end{tabular} 
\label{table:main_results}
\end{table*}

\begin{table*}[t]
\centering
\caption{Comparison with the best baseline method in Table~\ref{table:main_results} in terms of BLEU, BLEURT and YiSi. \ }
\begin{tabular}{l|lll|lll}
\toprule
\multirow{2}{*}{Methods}    &\multicolumn{3}{c|}{En$\rightarrow$De}&\multicolumn{3}{c}{En$\rightarrow$Fr}\\
     & BLEU &BLEURT  &YiSi &BLEU &BLEURT &YiSi 
     \\
\hline
Joint Training with BART &30.88 & 0.225 & 0.837 &44.18 & 0.488 & 0.864\\
$F_2$-XEnDec &\textbf{31.60} &\textbf{0.261} & \textbf{0.842} &\textbf{45.15} & \textbf{0.513} & \textbf{0.869}\\
\bottomrule
\end{tabular} \label{table:metrics}
\end{table*}
\subsection{Settings}
\textbf{Datasets.} We evaluate our approach on two representative, resource-rich translation datasets, WMT'14 English-German and WMT'14 English-French across four translation directions, English$\rightarrow$German (En$\rightarrow$De), German$\rightarrow$English (De$\rightarrow$En), English$\rightarrow$French (En$\rightarrow$Fr), and French$\rightarrow$English (Fr$\rightarrow$En). To fairly compare with previous state-of-the-art results on these two tasks, we report case-sensitive tokenized BLEU scores calculated by the {\em multi-bleu.perl} script.
The English-German and English-French datasets consist of 4.5M and 36M sentence pairs, respectively.
The English, German and French monolingual corpora in our experiments come from the WMT'14 translation tasks. We concatenate all the newscrawl07-13 data for English and German, and newscrawl07-14 for French which results in 90M English sentences, 89M German sentences, and 42M French sentences.
We use a word piece model~\citep{schuster2012japanese} to split tokenized words into sub-word units. For English-German, we build a shared vocabulary of 32K sub-words units. The validation set is newstest2013 and the test set is newstest2014. The vocabulary for the English-French dataset is also jointly split into 44K sub-word units. The concatenation of newstest2012 and newstest2013 is used as the validation set while newstest2014 is the test set. Refer to the supplementary document for more detailed data pre-processing.

\textbf{Model and Hyperparameters.}
We implement our approach on top of the Transformer model~\citep{Vaswani:17} using the {\em Lingvo} toolkit~\citep{shen2019lingvo}.
The Transformer models follow the original network settings~\citep{Vaswani:17}.
In particular, the layer normalization is applied after each residual connection rather than before each sub-layer.
The dropout ratios are set to $0.1$ for all Transformer models except for the Transformer-big model on English-German where $0.3$ is used. We search the hyperparameters using the Transformer-base model on English-German. In our method, the shuffling ratio $p_{1}$ is set to $0.50$ while $0.25$ is used for English-French in Table~\ref{table:back_translation}. $p_{2}$ is sampled from a Beta distribution $Beta(2, 6)$. The dropout ratio of $\mA$ is $0.2$ for all the models. For decoding, we use a beam size of $4$ and a length penalty of $0.6$ for English-German, and a beam size of $5$ and a length penalty of $1.0$ for English-French. We carry out our experiments on a cluster of $128$ P100 GPUs and update gradients synchronously.
The model is optimized with Adam~\citep{kingma2014adam} following the same learning rate schedule used in~\citep{Vaswani:17} except for {\em warmup\_steps} which is set to $4000$ for both Transform-base and Transformer-big models.  

\textbf{Training Efficiency.}
When training the vanilla Transformer model, each batch contains $4096\times128$ tokens of parallel sentences on a $128$ P100 GPUs cluster. 
As there are three losses included in our training objective (Eq.~\eqref{eq:loss_all}) and the inputs for each of them are different, 
we evenly spread the GPU memory budget into these three types of data by letting each batch include $2048\times128$ tokens. Thus the total batch size is $2048\times128\times3$. The training speed is on average about $60\%$ of the standard training speed. The additional computation cost is partially due to the implementation of the noise function to corrupt the monolingual sentence $\vy^u$ and can be reduced by caching noisy data in the data input pipeline. Then the training speed can accelerate to about $80\%$ of the standard training speed.
\subsection{Main Results}\label{sec:exp_main}
Table \ref{table:main_results} shows the main results on the English-German and English-French datasets.
Our method is compared with the following strong baseline methods.
\cite{ott2018scaling} is the scalable Transformer model. Our reproduced Transformer model performs comparably with their reported results.
\cite{Cheng:19} is a  NMT model with adversarial augmentation mechanisms in supervised learning.
~\cite{nguyen2019data} boosts NMT performance by adopting multiple rounds of back-translated sentences.
Both~\cite{zhu2020bertnmt} and~\cite{yang2019towards} incorporate the knowledge of pre-trained models into NMT models by treating them as frozen input representations for NMT.
We also compare our approach with MASS~\cite{song2019mass} and BART~\cite{lewis2019bart}. As their goals are to learn generic pre-trained representations from massive monolingual corpora, for fair comparisons, we re-implement their methods using the same backbone model as ours, and jointly optimize their self-supervised objectives together with the supervised objective on the same corpora.

For English-German, our approach achieves significant improvements in both translation directions over the standard Transformer model.
Even compared with the strongest baseline on English$\rightarrow$German, our approach obtains a $+0.72$ BLEU gain. More importantly, when we apply our approach to a significantly larger dataset, English-French with $36$M sentence pairs (vs.~English-German with $4.5$M sentence pairs), it still yields consistent and notable improvements over the standard Transformer model.

The single-stage approaches (Joint Training with MASS \& BART)  perform slightly better than the two-stage approaches \cite{zhu2020bertnmt,yang2019towards}, which substantiates the benefit of jointly training supervised and self-supervised objectives for resource-rich translation tasks. Among them, BART performs better with stable improvements on English-German and English-French and faster convergence. However, they still lag behind our approach. This is mainly because the $\mathcal{L}_{F_{2}}$ term in our approach can deeply fuse the supervised and self-supervised objectives instead of simply summing up their training losses. See Section~\ref{sec:ablation} for more details.

Furthermore, we evaluate our approach and the best baseline method (Joint Training with BART) in Table~\ref{table:metrics} 
in terms of two additional evaluation metric, BLEURT~\citep{sellam2020bleurt} and YiSi~\citep{lo2019yisi}, which claim better correlation with human judgement. Results in Table~\ref{table:metrics} corroborate the superior
performance of our approach compared to the best baseline method on both English$\rightarrow$German and English$\rightarrow$French. 


\subsection{Analyses}\label{sec:exp_analysis}
\begin{table}[t]
    \centering
    \caption{Effect of monolingual corpora sizes.}
    \begin{tabular}{lll}
    \toprule
    Methods &Mono. Size &En$\rightarrow$De
    \\ \midrule
    \multirow{5}{*}{\fancyname} &$\times 0$  &$28.70$ \\
    &$\times 1$  &$29.84$ \\
    &$\times 3$ &$30.36$ \\
    &$\times 5$ &$30.46$ \\
    &$\times 10$ &$30.22$ \\
    \bottomrule
    \end{tabular} \label{table:corpora_size}
\end{table}
\textbf{Effect of Monolingual Corpora Sizes.} Table~\ref{table:corpora_size} shows the impact of monolingual corpora sizes on the performance for our approach.
We find that our approach already yields improvements over the baselines when using no monolingual corpora (x0) as well as 
when using a monolingual corpus with size comparable to the bilingual corpus (1x). As we increase the size of the monolingual corpus to 5x, we obtain the best performance with $30.46$ BLEU. However, continuing to increase the data size fails to improve the performance any further. A recent study~\citep{liu2020very} shows that increasing the model capacity has great potential to exploit extremely large training sets for the Transformer model. We leave this line of exploration as future work.

\begin{table}[t]
        \centering
        \caption{Finetuning vs. Joint Training.}
        \begin{tabular}{l|l}
        \toprule
        Methods &En$\rightarrow$De
        \\ \midrule
        Transformer & $28.70$ \\
        + Pretrain + Finetune &$28.77$ \\
        \midrule
        \fancyname~(Joint Training)  &$30.46$ \\
        + Pretrain + Finetune &$29.70$ \\
        \bottomrule
        \end{tabular} \label{table:finetune}
\end{table}
\begin{table}[t]
\centering
\caption{Results on $\fancyname$~+ Back Translation. Experiments on English-German and English-French are based on the Transformer-big model.}
\begin{tabular}{l|ll}
\toprule
Methods &En$\rightarrow$De & En$\rightarrow$Fr\\ 
\midrule
Transformer & $28.70$  &$43.37$ \\
Back Translation &$32.09$ &$35.90$ \\
\cite{Edunov:18} &$\textbf{35.00}$\tablefootnote{Our results cannot directly
be compared to the numbers in~\citep{Edunov:18} because they use WMT’18 as bilingual data (5.18M) and 10x
more monolingual data (226M vs. ours 23M).} &$45.60$ \\
\midrule
\fancyname &$31.60$ &$45.15$ \\
+ Back Translation &$33.70$ &$\textbf{46.19}$ \\
\bottomrule
\end{tabular} \label{table:back_translation}
\end{table}
\textbf{Finetuning vs. Joint Training.} To further study the effect of pre-trained models on the Transformer model and our approach, we use Eq.~\eqref{eq:loss_self} to pre-train an NMT model on the entire English and German monolingual corpora. Then we finetune the pre-trained model on the parallel English-German corpus. 
Models finetuned on pre-trained models usually perform better than models trained from scratch at the early stage of training. However, this advantage gradually vanishes as training progresses~(cf.~Figure 1 in the supplementary document). 
As shown in Table~\ref{table:finetune}, Transformer with finetuning achieves virtually identical results as a Transformer trained from scratch.
Using the pre-trained model over our approach impairs performance. We believe this may be caused by a discrepancy between the pre-trained loss and our joint training loss. 

\textbf{Back Translation as Noise.}
One widely applicable method to leverage monolingual data in NMT is back translation~\citep{Sennrich:16b}. A straightforward way to incorporate back translation into our approach is to treat back-translated corpora as parallel corpora.
However, back translation can also be regarded as a type of noise used for constructing $\vy^{u}_{noise}$ in \fancyname~(shown in Fig.~\ref{figure:approach_b} and Step 4 in Algorithm~\ref{algo1}), which can increase the noise diversity. As shown in Table~\ref{table:back_translation}, for English$\rightarrow$German trained on the Transformer-big model, our approach yields an additional $+1.9$ BLEU gain when using back translation to noise $\vy^{u}$ and also outperforms the back-translation baseline.
When applied to the English-French dataset, we achieve a new state-of-the-art result over the best baseline~\citep{Edunov:18}. In contrast, the standard back translation for English-French hurts the performance of Transformer, which is consistent with what was found in previous works, \eg~\citep{caswell2019tagged}. These results show that our approach is complementary to the back-translation method and performs more robustly when back-translated corpora are less informative 
although our approach is conceptually different from works related to back translation~\citep{Sennrich:16b,Cheng:16,Edunov:18}. 

\textbf{Robustness to Noisy Inputs.}
Contemporary NMT systems often suffer from dramatic performance drops when they are exposed to input perturbations~\citep{Belinkov:17,Cheng:19}, even though these perturbations may not be strong enough to alter the meaning of the input sentence. In this experiment, we verify the robustness of the NMT models learned by our approach.
Following~\citep{Cheng:19}, we evaluate the model performance against word perturbations which specifically includes two types of noise to perturb the dataset.
The first type of noise is code-switching noise (CS) which randomly replaces words in the source sentences with their corresponding target-language words. Alignment matrices are employed to find the target-language words in the target sentences.
The other one is drop-words noise (DW) which randomly discards some words in the source sentences. Figure~\ref{figure:robustness} shows that our approach exhibits higher robustness than the standard Transformer model across all noise types and noise fractions. In particular, our approach performs much more stable for the code-switching noise.

\begin{figure}[t]
\centering
\includegraphics[width=0.48\textwidth]{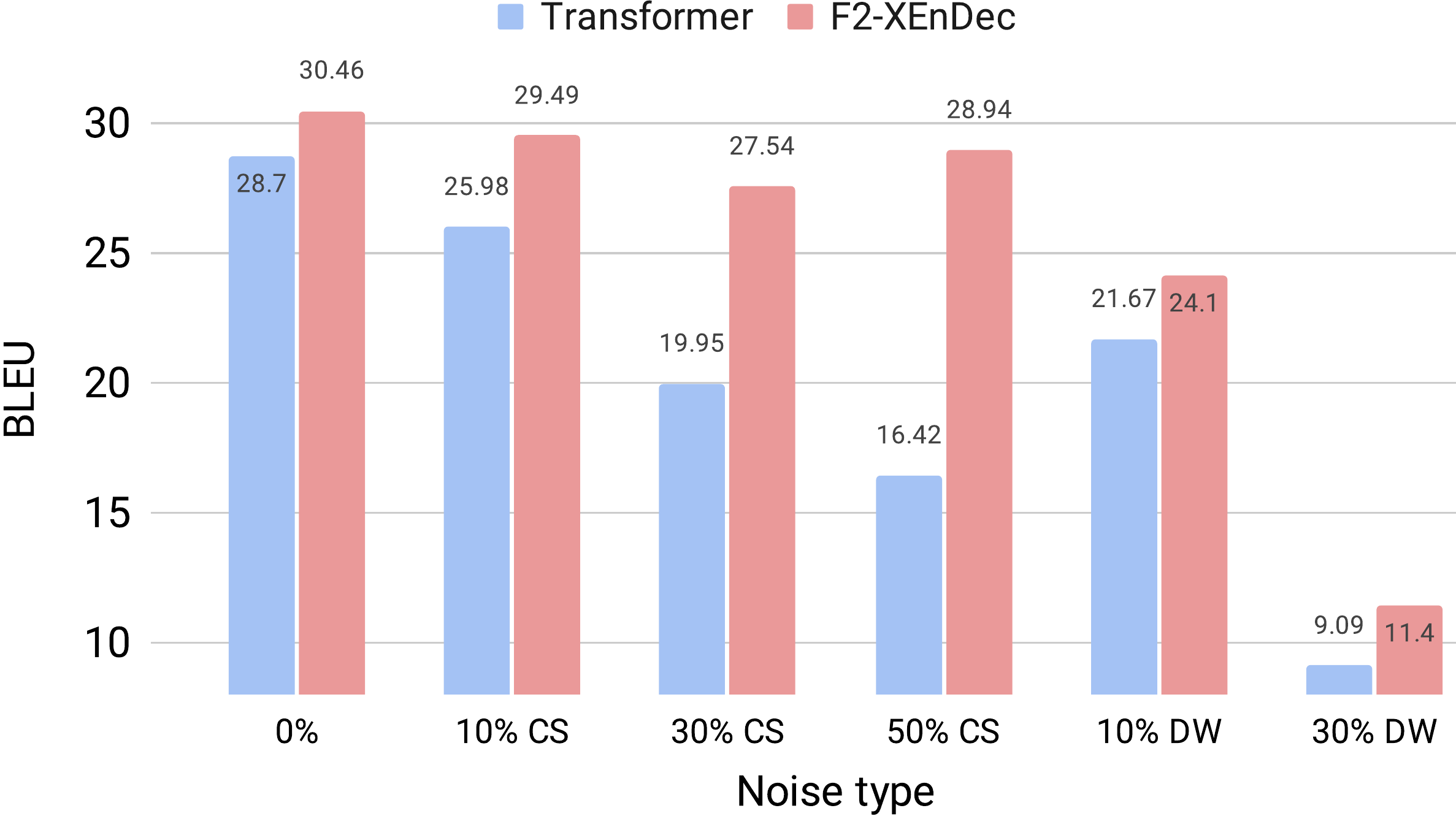} 
\caption{Results on artificial noisy inputs. ``CS": code-switching noise. ``DW'': drop-words noise. We compare our approach to the standard Transformer on different noise types and fractions.
}\label{figure:robustness}
\end{figure}

\begin{table}[t]
\begin{center}
\caption{Ablation study on English-German.}
\begin{tabular}{ll|l}

\toprule
ID & Different Settings &BLEU
\\ \midrule
1&  Transformer  &$28.70$ \\
\midrule
2& \fancyname &$30.46$ \\
3& without $\mathcal{L}_{F_2}$ &$29.21$ \\
4& without $\mathcal{L}_{F_1}$ (a prior alignment is used) &$29.55$ \\
5& $\mathcal{L}_{\mathcal{S}}$ with \emph{XEnDec} over parallel data &$29.23$ \\
6& \emph{XEnDec} is replaced by Mixup &$29.67$ \\
7& without dropout $\mA$ and model predictions  &$29.87$ \\
8& without model predictions &30.24 \\
 \bottomrule
\end{tabular} \label{table:ablation_study}
\end{center}
\end{table}

\subsection{Ablation Study}\label{sec:ablation}
Table~\ref{table:ablation_study} studies the contributions of the key components and verifies the design choices in our approach.

\noindent\textbf{Contribution of $\mathcal{L}_{F_2}$.} We first verify the importance of our core loss term $\mathcal{L}_{F_2}$. 
When $\mathcal{L}_{F_2}$ is removed in Eq.~\eqref{eq:loss_all}, the training objective is equivalent to summing up supervised ($\mathcal{L}_{S}$) and self-supervised ($\mathcal{L}_{F_1}$) losses. By comparing Row 2 and 3 in Table~\ref{table:ablation_study}, we observe a sharp drop (-1.25 BLEU points) caused by the absence of $\mathcal{L}_{F_2}$. This result demonstrates the crucial role of the proposed \emph{$F_2$-XEnDec} that can extract  complementary signals to facilitate the joint training. We believe this is 
because of the deep fusion of monolingual and parallel sentences at instance level.


\noindent\textbf{Inputs to \emph{XEnDec}.} To validate the proposed task,~\emph{XEnDec}, we apply it over different types of inputs. The first one directly combines the parallel and monolingual sentences without using noisy monolingual sentences, which is equivalent to removing $\mathcal{L}_{F_1}$ (Row 4 in Table~\ref{table:ablation_study}). We achieve this by setting $n(\vy^{u}) =\vy^{u} $ in Algorithm~\ref{algo1}. 
However, we cannot obtain $\mA^{\prime}$ required by the algorithm (Line 7 in Algorithm~\ref{algo1}) which leads to the failure of calculating the loss. Thus we design a prior alignment matrix to handle this issue (cf. Section 2 in the supplementary document). The second experiment utilizes \emph{XEnDec} only over parallel sentences (Row 5 in Table~\ref{table:ablation_study}). We can find these two cases can both achieve better performance compared to the standard Transformer model (Row 1 in  the table). These results show the efficacy of the proposed \emph{XEnDec} on different types of inputs. Their gap to our final method shows the rationale of using both parallel and monolingual sentences as inputs. We hypothesize this is because \emph{XEnDec} implicitly regularizes the model by shuffling and reconstructing words in parallel and monolingual sentences.


\noindent\textbf{Comparison to Mixup.} 
We use Mixup~\cite{Zhang:18} to replace our second \emph{XEnDec} to compute $\mathcal{L}_{F_{2}}$ while keeping the first \emph{XEnDec} untouched. When applying Mixup on a pair of training data $(\vx, \vy)$ and $(\vxp, \vyp)$, Eq.~\eqref{eq:mix_src}, Eq.~\eqref{eq:mix_tgt} and Eq.~\eqref{eq:mix_signal} are replaced by $e(\tilde{x}_{i}) = \lambda e(x_i) + (1-\lambda) e(x^{\prime}_i)$, $e(\tilde{z}_{j}) = \lambda e(y_{j-1}) + (1 - \lambda) e(y^{\prime}_{j-1})$ and $h(\tilde{y_j}) =  \lambda h(y_j) + (1-\lambda) h(y^{\prime}_j)$, respectively, where $\lambda$ is sampled from a Beta distribution.
The comparison between Row 6 and Row 2 in Table~\ref{table:ablation_study} shows that Mixup leads to a worse result. Different from Mixup which encourages the model to behave linearly to the linear interpolation of training examples, our task combines training examples in a non-linear way in the source end, and forces the model to decouple the non-linear integration in the target end while predicting the entire target sentence based on its partial source sentence. 

\noindent\textbf{Computation of $\mA$ and $h(\tvy)$.} The last two rows in  Table~\ref{table:ablation_study} verify the impact of two training techniques discussed in Section~\ref{sec:our_algorithm}. Removing these components would lower the performance.

\eat{
\begin{figure}[h]
\begin{center}
\includegraphics[width=0.50\textwidth]{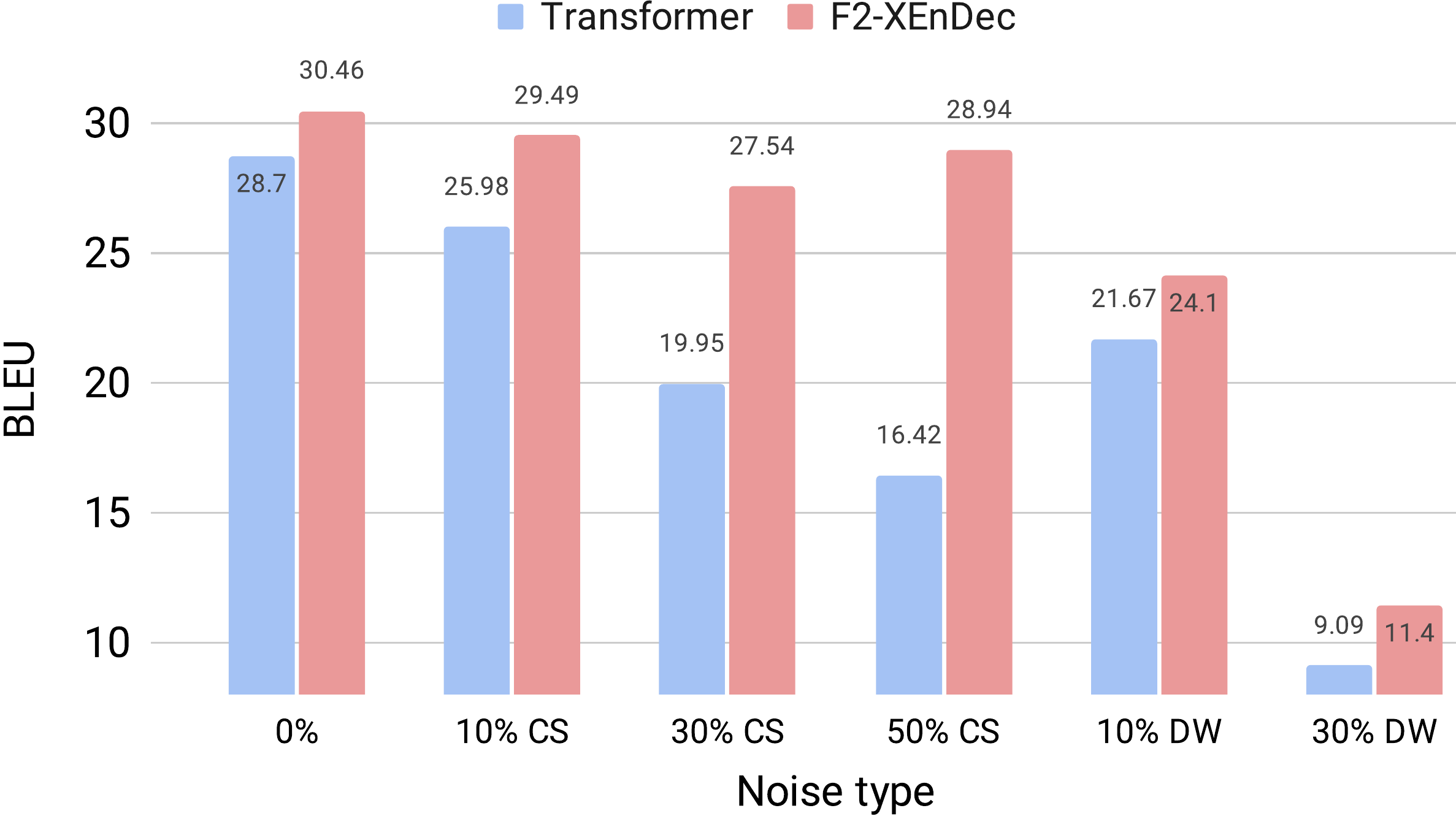} 
\caption{Results on artificial noisy inputs. ``CS" and ``DW'' stands for code-switched noise and drop-words noise respectively. We compare our approach to the standard Transformer on different noise types and fractions.}
\label{figure:robustness}
\end{center}
\end{figure}
}

\section{Related Work}
The recent past has witnessed an increasing interest in the research community on leveraging pre-training models to boost NMT model performance~\citep{ramachandran2016unsupervised, lample2019cross, song2019mass, lewis2019bart, Edunov:18, zhu2020bertnmt, yang2019towards,liu2020multilingual}. Most successes come from low-resource and zero-resource translation tasks. \cite{zhu2020bertnmt} and \cite{yang2019towards} achieve some promising results on resource-rich translations. They propose to combine NMT model representations and frozen pre-trained representations under the common two-stage framework. The bottleneck of these methods is that these two stages are decoupled and separately learned, which exacerbates the difficulty of finetuning self-supervised representations on resource-rich language pairs. Our method, on the other hand, jointly trains self-supervised and supervised NMT models to close the gap between representations learned from either of them with an essential new subtask, \emph{XEnDec}. In addition, our new subtask can be applied to combine different types of inputs.
Experimental results show that our method consistently outperforms previous approaches across several translation benchmarks and establishes a new state-of-the-art result on WMT'14 English-French when applying \emph{XEnDec} to back-translated corpora.

Another line of research related to ours originates in computer vision by interpolating images and their labels~\citep{Zhang:18,yun2019cutmix} which have been shown effective in improving generalization~\cite{arazo2019unsupervised,jiang2020beyond,xu2021faster,northcutt2021confident} and robustness of convolutional neural network~\citep{hendrycks2019augmix}. Recently, some research efforts have been devoted to introducing this idea to NLP applications~\cite{cheng2020advaug,guo:sequencemixed,chen2020mixtext}.
Our \emph{XEnDec} shares the commonality of combining example pairs. However, \emph{XEnDec}'s focus is on sequence-to-sequence learning for NLP with the aim of using self-supervised learning to complement supervised learning in joint training.

\section{Conclusion}
This paper has presented a joint training approach, \fancyname, to combine self-supervised and supervised learning in a single stage.
The key part is a novel cross encoder-decoder which can be used to ``interbreed'' monolingual and parallel sentences, which can also be fed with different types of inputs and recover some popular self-supervised and supervised training objectives.

Experiments on two resource-rich translation tasks, WMT'14 English-German and WMT'14 English-French, show that joint training performs favorably against two-stage training approaches when an enormous amount of labeled and unlabeled data is available. When applying \emph{XEnDec} to deeply fuse monolingual and parallel sentences resulting in \fancyname, the joint training paradigm can better exploit the complementary signal from unlabeled data with significantly stronger performance. Finally, \fancyname~is capable of improving the NMT robustness against input perturbations such as code-switching noise widely found in social media. 

In the future, we plan to further examine the effectiveness of our approach on larger-scale corpora with high-capacity models. We also plan to design more expressive noise functions for our approach.

\section*{Acknowledgements}
The authors would like to thank anonymous reviewers for
insightful comments, Isaac Caswell for providing back-translated corpora, and helpful feedback for the early version of this paper.
\bibliography{icml2021}
\bibliographystyle{icml2021}

\newpage
\appendix
\section*{Appendix}
\begin{figure}[ht]
	\centering
	\begin{subfigure}{}
		\includegraphics[width=6cm,height=5cm]{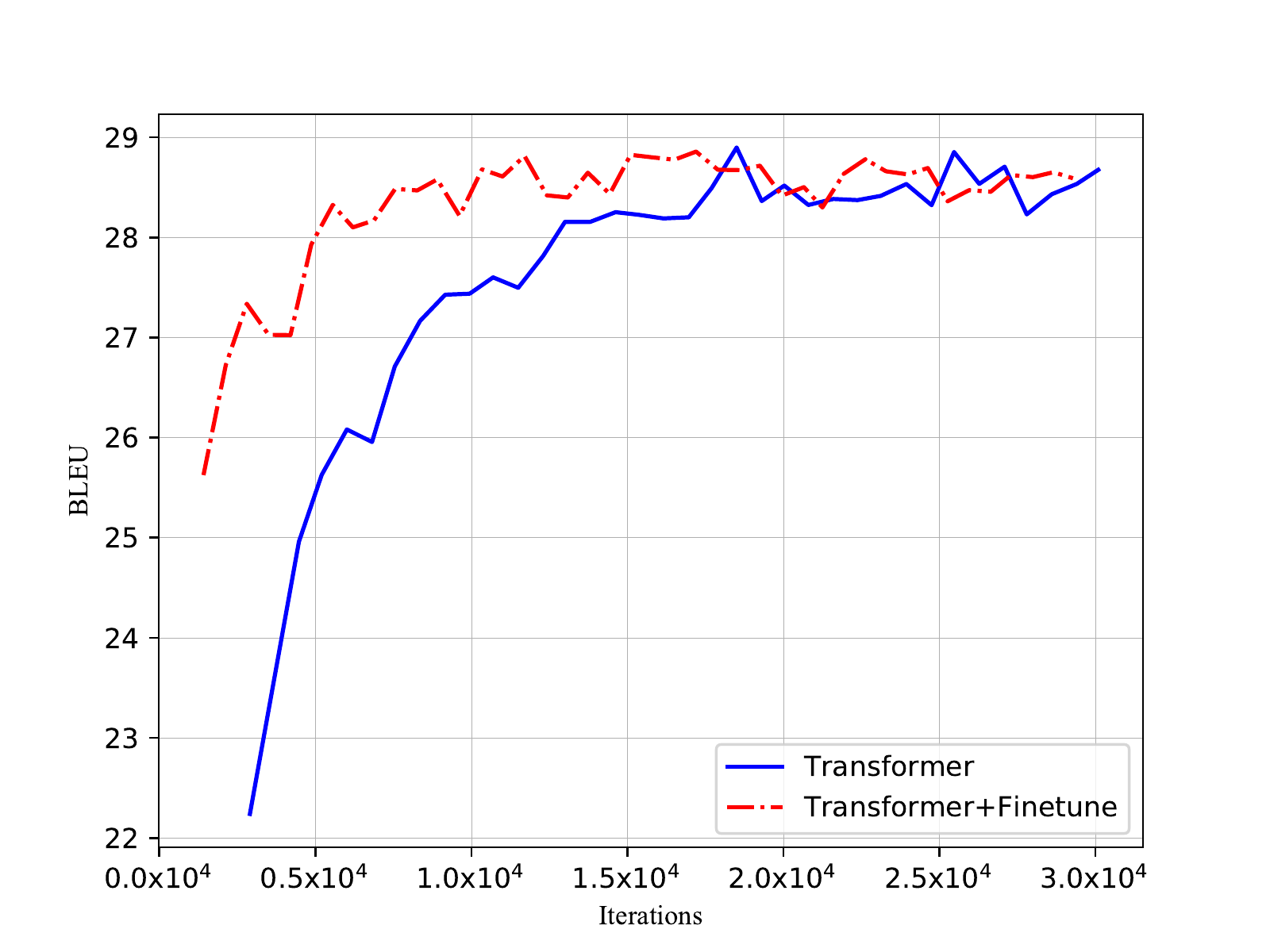}
	\end{subfigure}
	\hfill
	\begin{subfigure}{}
		\includegraphics[width=6cm,height=5cm]{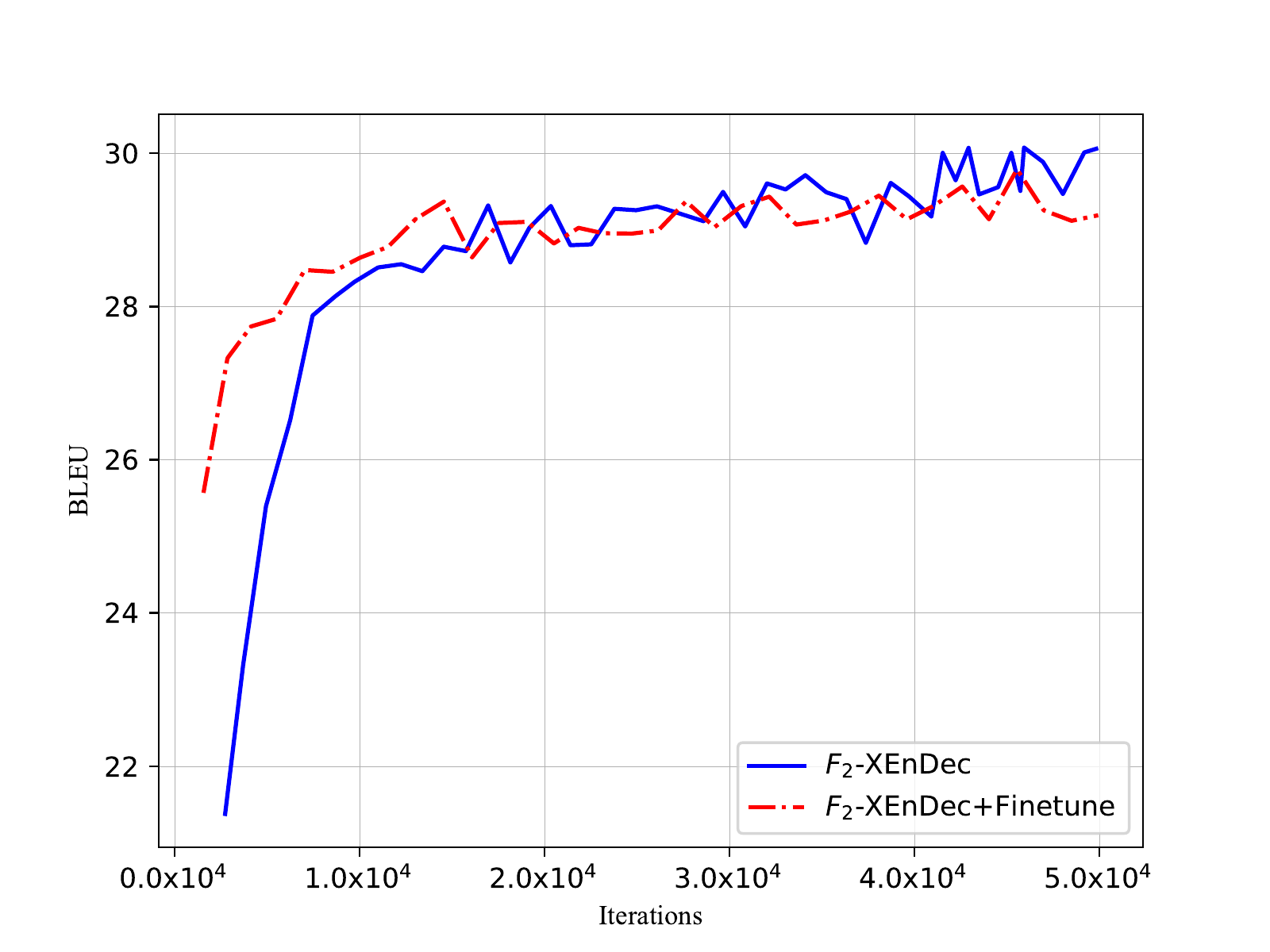}
	\end{subfigure}
		\caption{Comparison of finetuning and training from scratch using Transformer and \fancyname. In both methods, pre-training leads to faster convergence but fails to improve the final performance after the convergence. The comparison between the figures shows our joint training approach on the left (the blue curve) significantly outperforms against the two-stage training on the right. Final BLEU numbers are reported in Table 5 in the main paper.  }\label{figure:finetune}
\end{figure}

\section{Training Details} \label{traingdetails}
\textbf{Data Pre-processing} We mainly follows the pre-processing pipeline \footnote{\url{https://github.com/pytorch/fairseq/tree/master/examples/translation}} which is also adopted by~\cite{ott2018scaling}, ~\cite{Edunov:18} and~\cite{zhu2020bertnmt}, except for the sub-word tool. To verify the consistency between the word piece model~\citep{schuster2012japanese} and the BPE model~\citep{Sennrich:16a}, we conduct a comparison experiment to train two standard Transformer models using the same data set processed by the word piece model and the BPE model respectively. The BLEU difference between them is about ±0.2, which suggests there is no significant difference between them.

\textbf{Batching Data} Transformer groups training examples of similar lengths together with a varying batch size for training efficiency ~\citep{Vaswani:17}. In our approach, when interpolating two source sentences, $\bx^{p}$ and $\vy^{\diamond}$, it is better if the lengths of $\bx^{p}$ and $\vy^{\diamond}$ are similar, which can reduce the chance of wasting positions over padding tokens.
To this end, in the first round, we search for monolingual sentences with exactly the same length of the source sentence in a parallel sentence pair. After the first traversal of the entire parallel data set, we relax the length difference to $1$.
This process is repeated by relaxing the constraint until all the parallel data are paired with their own monolingual data.

\section{A Prior Alignment Matrix} ~\label{prior_alignment}
When $\mathcal{L}_{F_1}$ is removed, we can not obtain $\mA^{\prime}$ according to Algorithm 1 in the main paper which leads to the failure of calculating $\mathcal{L}_{F_2}$. Thus we propose a prior alignment to tackle this issue. For simplicity, we set $n(\cdot)$ to be a copy function when doing the first \emph{XEnDec}, which means that we just randomly mask some words in the first round of \emph{XEnDec}. In the second \emph{XEnDec}, we want to combine $(\vx^{p}, \vy^{p})$ and $(\vy^{\diamond}, \vy)$. The alignment matrix $\mA^{\prime}$ for $(\vy^{\diamond}, \vy)$ is constructed as follows.

If a word $y_{j}$ in the target sentence $\vy$ is picked in the source side which indicates $y_{j}^{\diamond}$ is picked and $m_{j} = 0$, its attention value $A_{ji}^{\prime}$ if $m_{i}=0$ is assigned to $\frac{p}{\|1-\vm\|_1}$, otherwise it is assigned to $\frac{1 - p}{\|\vm\|_1}$ if $m_{i}=1$. Conversely, If a word $y_{j}$ is not picked which indicates $m_{j}=1$, its attention value $A_{ji}^{\prime}$ is assigned to $\frac{p}{\|\vm\|_1}$ if $m_{i}=0$, otherwise it is $\frac{1 - p}{\|1 - \vm\|_1}$ if $m_{i}=1$.
\balance
\end{document}